# Tracking, exploring and analyzing recent developments in German-language online press in the face of the coronavirus crisis:

cOWIDplus Analysis and cOWIDplus Viewer

Sascha Wolfer[1], Alexander Koplenig[1], Frank Michaelis[1], Carolin Müller-Spitzer[1]
[1]Leibniz Institute for the German Language, Mannheim, Germany

The coronavirus pandemic may be the largest crisis the world has had to face since World War II. It does not come as a surprise that it is also having an impact on language as our primary communication tool. We present three inter-connected resources that are designed to capture and illustrate these effects on a subset of the German language: An RSS corpus of German-language newsfeeds (with freely available untruncated unigram frequency lists), a static but continuously updated HTML page tracking the diversity of the used vocabulary and a web application that enables other researchers and the broader public to explore these effects without any or with little knowledge of corpus representation/exploration or statistical analyses.

## 1 The coronavirus pandemic and its influence on language

All around the globe, the coronavirus pandemic is affecting almost every aspect of public life. Consequently, the pandemic is *the* subject of discussion, not only in private face-to-face conversation (if you still have the opportunity to talk to someone face-to-face at all) but also in the news. With many activities of daily life like sports and cultural events coming to a halt, respective newspaper desks might very well run out of events to report on or shift their focus to pandemic-related topics. Other, more general desks like politics are also mainly dealing with the effects of the pandemic on society, for example, the impending crisis in the health-care system, curfews or other measures taken by national governments.

This leads to the assumption that the vocabulary used in articles, not only in printed but also in online media, is changing. To be precise, we assume a (temporary) frequency rise for specific word forms that are associated with the all-encompassing coronavirus pandemic. This does not exclusively mean that we can observe word forms that have never been observed in every-day press language. Although some terminological vocabulary (esp. from epidemiology and virology) might find its way into every-day press language, it is also very likely that concepts from every-day language which usually play a secondary role are suddenly much more important.

To investigate all these effects of the coronavirus pandemic on German press language, we created three inter-connected resources which we will introduce below: a corpus of RSS feeds, a dynamic HTML site which presents further analyses on vocabulary diversity throughout the coronavirus pandemic ("cOWIDplus Analysis") and an online viewer application that allows exploration of frequency profiles through time ("cOWIDplus Viewer"). The names of the two resources are derived from "OWIDplus" (www.owid.de/plus), which is an experimental platform for multilingual lexical-lexicographic data, for quantitative lexical analyses and for interactive lexical applications within the "Online-Wortschatz-Informationssystem Deutsch" provided by the Leibniz Institute for the German Language in Mannheim (OWID, www.owid.de).

## 2 RSS corpus

Since the beginning of 2020, we have been collecting a corpus of RSS feeds for 13 sources in German language: Focus Online (https://www.focus.de/), Frankfurter Allgemeine Zeitung (https://www.faz.net/), Frankfurter Rundschau (https://www.fr.de/), Süddeutsche Zeitung (https://www.sueddeutsche.de/), Neue Zürcher Zeitung (https://www.nzz.ch/), Spiegel Online (https://www.spiegel.de/), Der Standard (https://www.derstandard.at), tageszeitung (https://taz.de/), Die Welt (https://www.welt.de/), Die Zeit (https://www.zeit.de/). In addition to these more "traditional" news outlets, we included three outlets from editorial offices that have no counterpart in printed form with the same name: web.de, t-online.de, and heise.de. The sources were selected for popularity, variety in German-speaking countries (one source each from Austria and Switzerland), and political orientation (for example, the Frankfurter Allgemeine Zeitung is described as a conservative newspaper whereas the tageszeitung is known to have a left-wing orientation. Also, heise.de, a source with a strong focus on information technology, is included in the sample. We use a custom R (R Core Team, 2020) script and the XML package (Temple Lang, 2020) to collect the raw feed data. The RSS feeds are extracted every three hours. This leads to duplicated items in the collected data because some items can and will still be in the RSS feed after three hours. We exclude these items by only considering unique entries in the data structure. Note that this procedure could still leave “duplicates” in our data because sometimes the editorial staff decide to make changes to already published content (e.g., a small change in the title of an article). Technically, however, these count as new items in the source’s RSS feeds and are therefore treated as separate entries in our corpus.

## 2.1 Corpus preparation

All titles and descriptions (short pieces of text that introduce the articles) are extracted from the sources’ RSS feeds along with the associated publication timestamp. Due to differing timestamp formats, we have to normalize the timestamps using the R package lubridate (Grolemund & Wickham, 2011). We discard the exact timestamp and only keep the day of publication. We exclude all punctuation from the titles and description (henceforth “texts”) except hyphens because we want to keep compounds. HTML markup (e.g. <strong> or </strong>) that could be used in RSS feeds is also excluded from the texts. All characters are converted to lower-case.

A few specific items are excluded from the texts, mainly because they are very specific to certain sources. The following list is the comprehensive set of word forms that are excluded following the above-mentioned procedure: *t-onlinede-redakteurin*, *t-onlinede-redakteur* (*redakteur* is German for “editor”, *redakteurin* is the female form), *sport-live-blog*, *t-onlinede*, *focus-online-redakteurin*, *focus-online-redakteur*, *focus-online-reporter*, *spiegel-titelstory*, *faz-sprinter*, *heise*, *derstandardat*, *km/h*. Also, all YouTube links and digit-only wordforms are excluded. After these pre-processing steps, unigram and bigram frequency lists are created for each day.

## 2.2 Corpus size

Given the continuous nature of corpus collection, we can only provide a snapshot of the current (sub)corpus sizes. The following figures reflect the state of the corpus up to May 21st, 2020. Currently, the RSS corpus contains 10,455,026 tokens which are distributed over 279,791 types. The corpus sizes for each month are summarized in Table 1. Figure 1 breaks token sizes down to days. A linear model (the grey regression line is included in Figure 1) does not show a significant effect of the date on the size of the corpus on that day ($\beta = -12.75$, $SE = 31.56$, $t = -0.404$, $p = 0.687$).

Table 1: Monthly corpus measures for the RSS corpus. Please note that the figures for May only include May 1st to 21st.

| Month | Number of tokens | Share of tokens | Number of types |
|---|---|---|---|
| January 2020 | 2,240,164 | 22.6 % | 112,618 |
| February 2020 | 2,195,831 | 21.5 % | 108,256 |
| March 2020 | 2,309,229 | 22.1 % | 102,483 |
| April 2020 | 2,107,192 | 20.2 % | 96,911 |
| May 2020 (only till 21st) | 1,550,071 | 14.8 % | 82,862 |

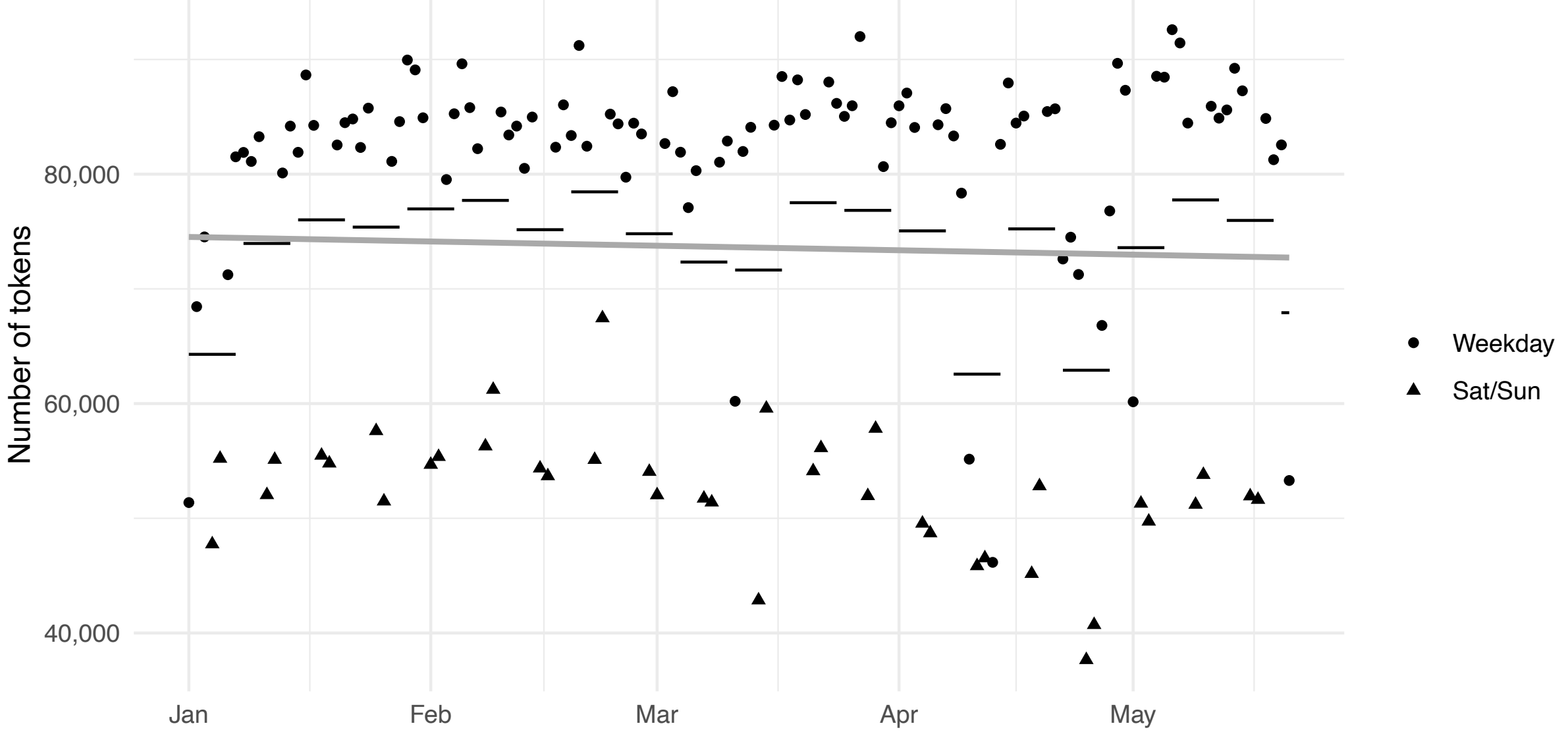

Figure 1: Daily corpus sizes of the RSS corpus with a linear model fit. Shapes indicate weekday or weekend (not taking into account holidays that did not fall on a weekend, e.g. May 1st). Horizontal line segments indicate the weekly mean (with day 1 of week 1 being January 1st, 2020).

Given these figures, we conclude that the size of the corpus can be thought of being more or less constant over time. To account for minor fluctuations (especially weekends vs. weekdays), we decided to report relative frequencies in the cOWIDplus Viewer (see Section 4).

## 3 cOWIDplus Analysis

With cOWIDplus Analysis (www.owid.de/plus/cowidplus2020), we provide a continuously updated resource in the form of an RMarkdown (Xie et al., 2018) HTML document that summarizes the findings with respect to the following research question: is there a quantitatively measurable narrowing of topics (and hence of the vocabulary) during the coronavirus pandemic in selected (online) news outlets published in German language? And if so, are topics (and hence the vocabulary) expanded after the crisis is "resolved"? Connected to this research question, we are testing two hypotheses: (1) Yes, there is indeed a narrowing of topics that can be measured by rather straightforward quantitative measures. A clear narrowing of vocabulary is indicated by all of the measures and by the investigation of frequency lists. (2) When the pandemic and its consequences are under control, the situation will normalize and the vocabulary in online news outlets will diversify again in a continuous way. The variables used to quantify vocabulary diversity will return to the pre-pandemic level.

The quantitative measures we use to operationalize the narrowing of the vocabulary are information-theoretic redundancy (Shannon, 1948: 14), mean segmental type-token ratio (MSTTR, Johnson, 1944: 3) and the accumulated token frequency share of the 100 most frequent word forms per day. The central idea of cOWIDplus Analysis is to update the resource each week with the language data up to this day and continuously evaluate the hypotheses. To enable the scientific public to replicate the analyses and to investigate further research questions, the following data is available for download on the cOWIDplus Analysis website: (1) Daily frequency lists (currently only of unigrams), (2) weekly frequency lists (of unigrams) and (3) the daily values for the central measures mentioned above (redundancy, MSTTR, and top 100 frequency share). This data is also updated weekly.

Since cOWIDplus Analysis is only available in German, we will not go into more detail regarding the methods and the current results. More information is available on the website.

## 4 cOWIDplus Viewer

Providing the daily frequency lists (see previous section) is a first step to enabling other researchers to use the RSS feed data. However, the amounts of knowledge and resources that are necessary to deal with the data could pose an obstacle for many people who might also be interested in the data. Inspired by the popular Google Books Ngram Viewer (Michel et al., 2011) we created the cOWIDplus Viewer (https://www.owid.de/plus/cowidplusviewer2020/) which enables users without any programming knowledge to search the RSS corpus for specific words forms, strings within word forms or bigrams.

## 4.1 Architecture and interface

The Viewer is a web application built using the Shiny framework (Chang et al., 2020). Shiny is an R package that allows to build interactive web apps from R. It comes with pre-defined HTML widgets that can be used to build an interactive browser-based user interface. On the server side, user input is processed within R and the results are returned to the user's browser.

The interface of the cOWIDplus Viewer is divided into two pages accessible via tabs at just below the title: the main page and the bigram finder page. Figure 2 shows a screenshot of the current English version of the cOWIDplus Viewer's main page.

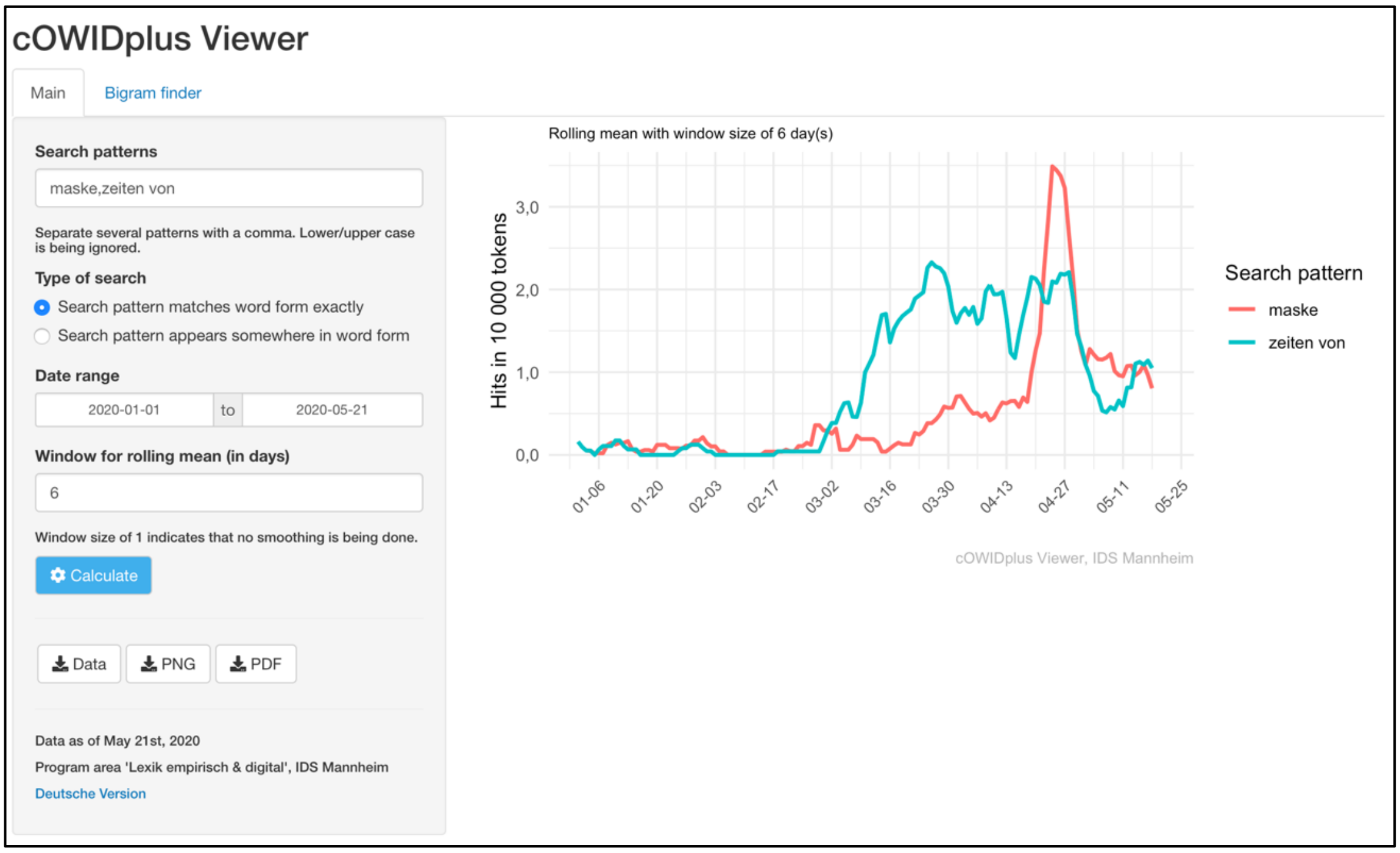

Figure 2: English version of the cOWIDplus Viewer main page with the default search patterns and the second search type.

All search parameters are located in the input panel on the left. Right at the top is the search pattern field where different patterns have to be separated by a comma. After experimenting with the possibility of entering regular expressions in the search field, we decided against this option because researchers who want to use regular expressions could still use the downloadable data set (see Section 3). In this case, we wanted to keep the entry threshold as low as possible and opted against regular expressions in the search field (which is, as a side note, also preferable due to security and, for some regular expressions, performance reasons). In fact, all special characters used in regular expressions are deleted from the users' inputs and white space around entries is trimmed before further processing. Instead of full-blown regular expressions, we opted for a simple switch where users can choose the type of search for unigram searches (bigrams are always searched exactly as entered by the user). The two options are exact matching (the internal search algorithm uses the regular expression ^<pattern>$ for this) and matching within other word forms. Furthermore, the user can choose a specific date range which is propagated to the x-axis of the frequency graph on the right, the table of absolute frequencies on the bottom and the downloadable data file. The last parameter sets the window size of the rolling mean calculation which is realized by the "frollmean" function (with a center-aligned window) in the R package data.table (Dowle & Srinivason, 2019) and can be chosen between 1 (no smoothing) and 14 days. Due to the potentially lengthy calculations, we chose to use an action button ("Calculate") to trigger the calculation. Normally, Shiny applications are "reactive", i.e. start calculation whenever an input parameter is changed. Three download buttons give the user the opportunity to download the data which is currently displayed (a CSV file with daily absolute frequencies, relative frequencies and smoothed values), and two types of graphic formats of the plot (a 250-dpi PNG and a vectorized PDF). Each week, the underlying corpus is updated, so we have to make sure that the user is aware of the current date of the data.

On the right side of the main page, we show a graph of the relative frequencies in time which is smoothed if a window of greater than 1 day is selected for the rolling mean. If the user selects the search type "within-word-form", a table is displayed below the plot. This table shows all hits for all patterns, sorted by absolute frequencies within the selected date range. This table is quite important for this type of search because it might well be the case that the word forms that actually appear most often do not match the search patterns exactly. This is also the case in Figure 2. The five most frequent word forms that are shown in the screenshot belong to the search pattern *maske* ("mask"). However, none of those hits is *maske* itself. There is *maskenpflicht* ("obligation to wear a mask"), *masken* (inflected form of *maske*), *schutzmasken* ("protective masks"), *masked* (part of the name of the TV show "The Masked Singer"), and *atemschutzmasken* ("respiratory masks"). The entry *maske* itself is ranked 7th in terms of absolute frequencies with 510 hits. The table can be searched, sorted and the users can flip pages (not visible in Figure 2).

Figure 3 shows the second page of the cOWIDplus Viewer (the "Bigram finder") with the default search parameters. We implemented the bigram finder with a potential user in mind who might not have a very clear-cut research question regarding a very specific bigram. With the bigram finder, this user can search the corpus to discover bigrams that might be worth checking out on the main page. There are three types of bigram searches: the first (and default) one searches for the pattern somewhere in a bigram, partial matches are allowed. The second and third types are more restrictive: here, one can search for bigrams where the search pattern is (exactly) the first (second option) or second part (third option) of the bigram. Consequently, the default search pattern *corona* returns bigrams like *das coronavirus* for the first option but only patterns like *corona und* ("corona and") for the second option or *wegen corona* ("because of corona") for the third option.

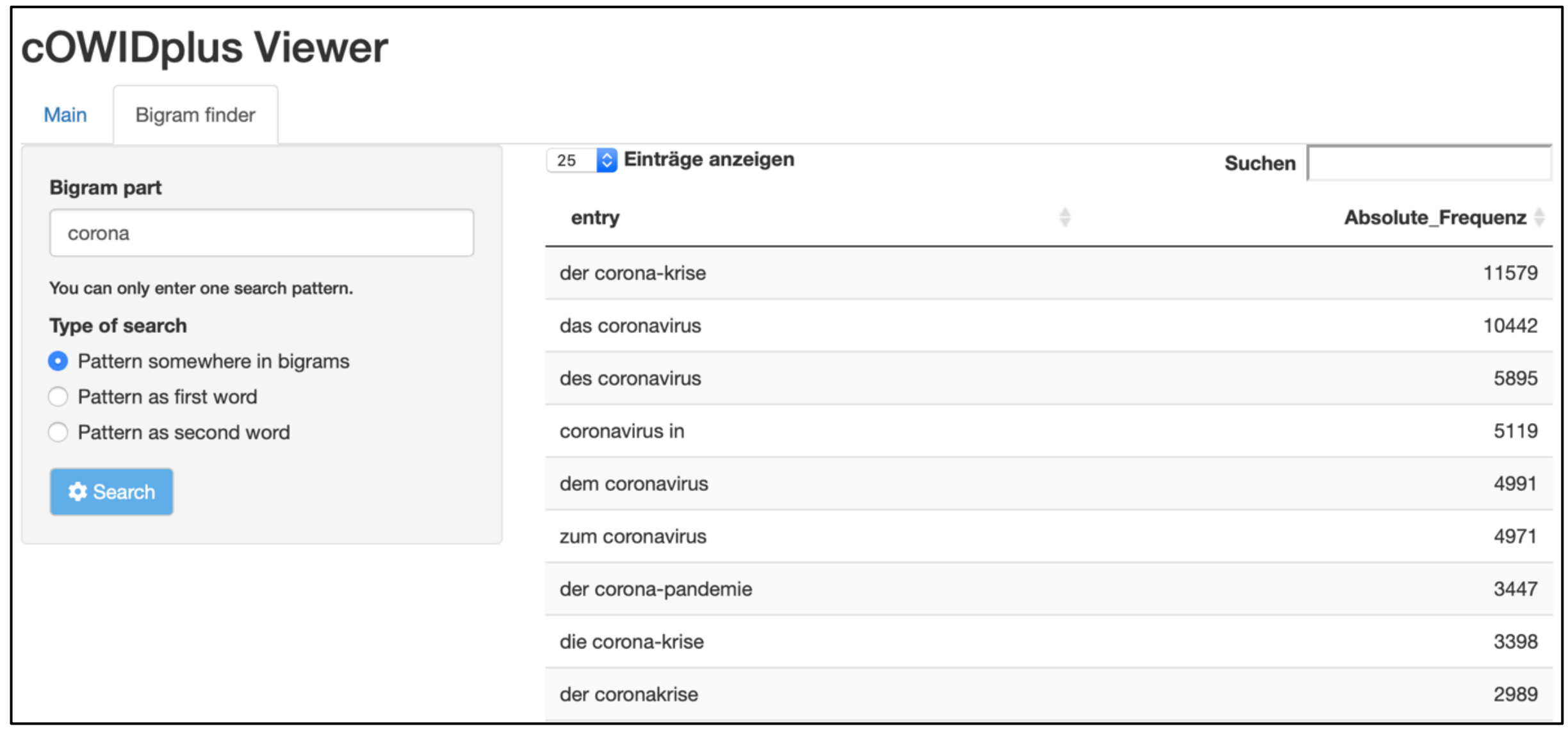

Figure 3: The "Bigram finder" of the cOWIDplus Viewer (second page of the application).

### 4.2 Examples

At the moment, most of the effects people might be interested in are the effects related to the coronavirus pandemic, as already introduced in Section 1. Here, we show query examples where effects of the pandemic on social life are quite evident. One such pair of queries extracted with a method outlined in Koplenig (2017) is *fc* (abbreviation for *Fußballclub*, "football club") and *corona*. These two word forms show almost perfect opposite frequency time courses. As soon as all games in the Bundesliga (Germany's men's football competition) were cancelled (the last fixture in the first division was played on March 11th, 2020), the relative frequency of *fc* starts to drop dramatically with *corona* gaining substantially. With the Bundesliga resuming on May 16th, the two patterns are likely to converge in relative frequencies again. Note, that only exact matches to *corona* are counted here. The effect would be even more pronounced if we would include all word forms where *corona* is included (esp. *coronavirus*).

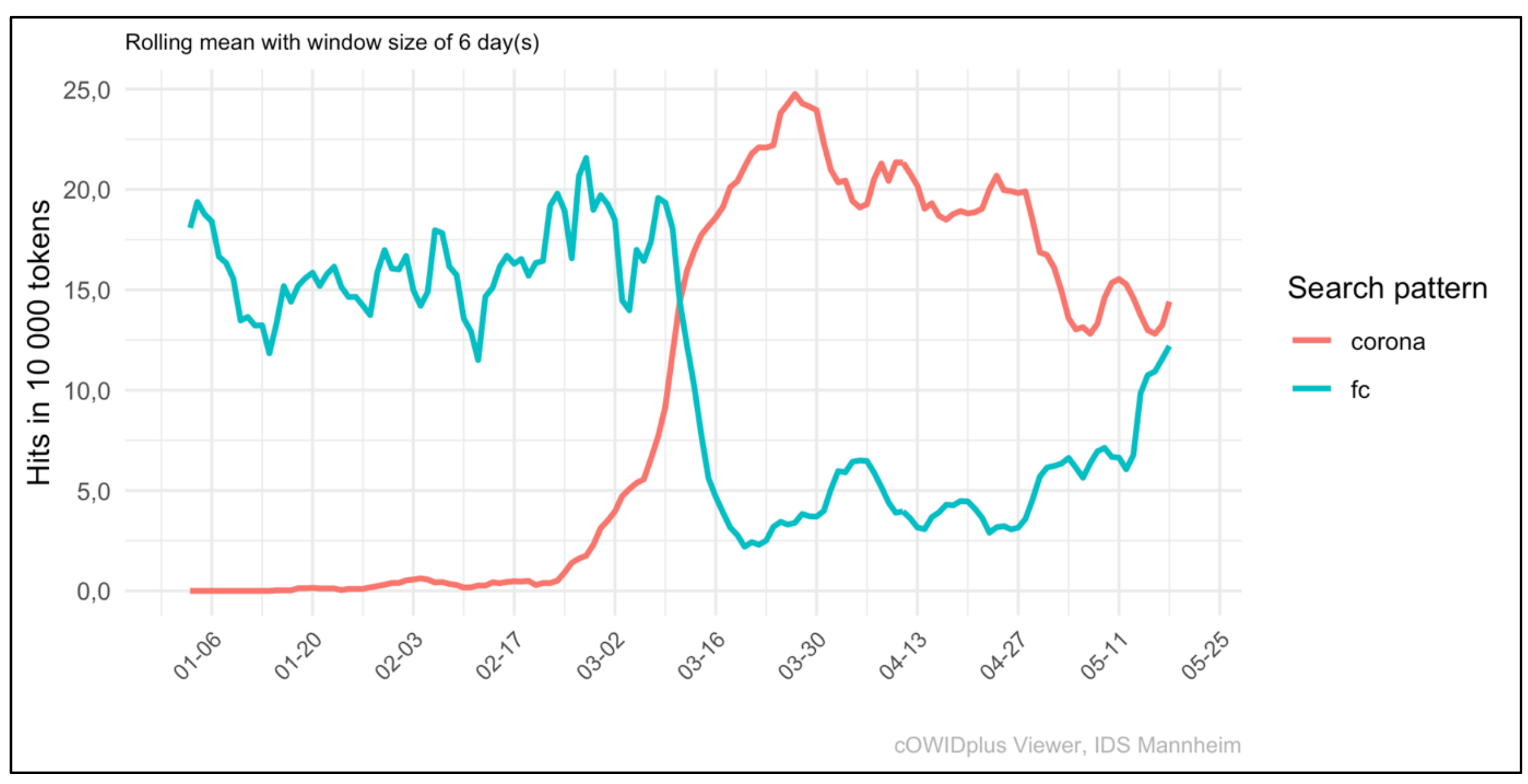

Figure 4: cOWIDplus Viewer result graph for exact matches of the word forms *fc* and *corona*.

Another example gives an impression of both a linguistically interesting case of naming conventions combined with a glimpse of political discussions. Figure 5 shows the developments in relative frequencies for the (partial) matches of *ausgangssperre* (curfew), *kontaktverbot* ("contact ban"), and *lockerung* ("relaxation"). The first mention of a potential curfew captured in the RSS corpus was on March 12th. Afterwards, its frequency peaked on March 15th. However, a curfew was never implemented nation-wide in Germany. Instead, on March 22nd, it was decided to implement a contact ban (which could be described as a much lighter version of a curfew).

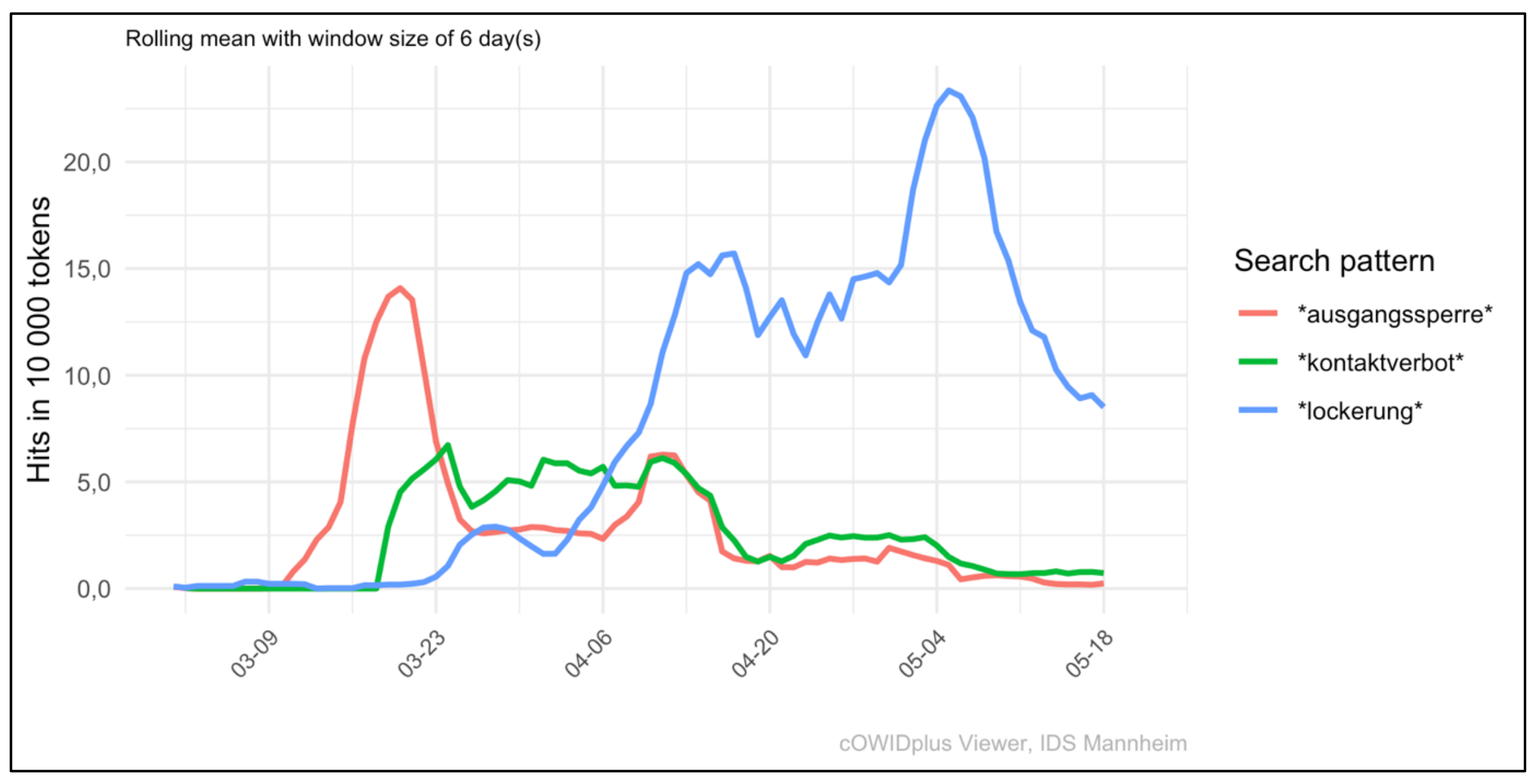

Figure 5: cOWIDplus Viewer result graph for matches (also in other word forms) of *ausgangssperre* ("curfew"), *kontaktverbot* ("contact ban"), and *lockerung* ("relaxation").

Although *kontaktverbot* never reached a peak as high as *ausgangssperre*, it was consistently more frequent than *ausgangssperre* for a prolonged time. Then, around mid-April, discussions about potential relaxations began which is reflected by a sharp increase of frequencies of *lockerung*. With certain relaxations becoming reality at the beginning of May, the frequencies increased even more.

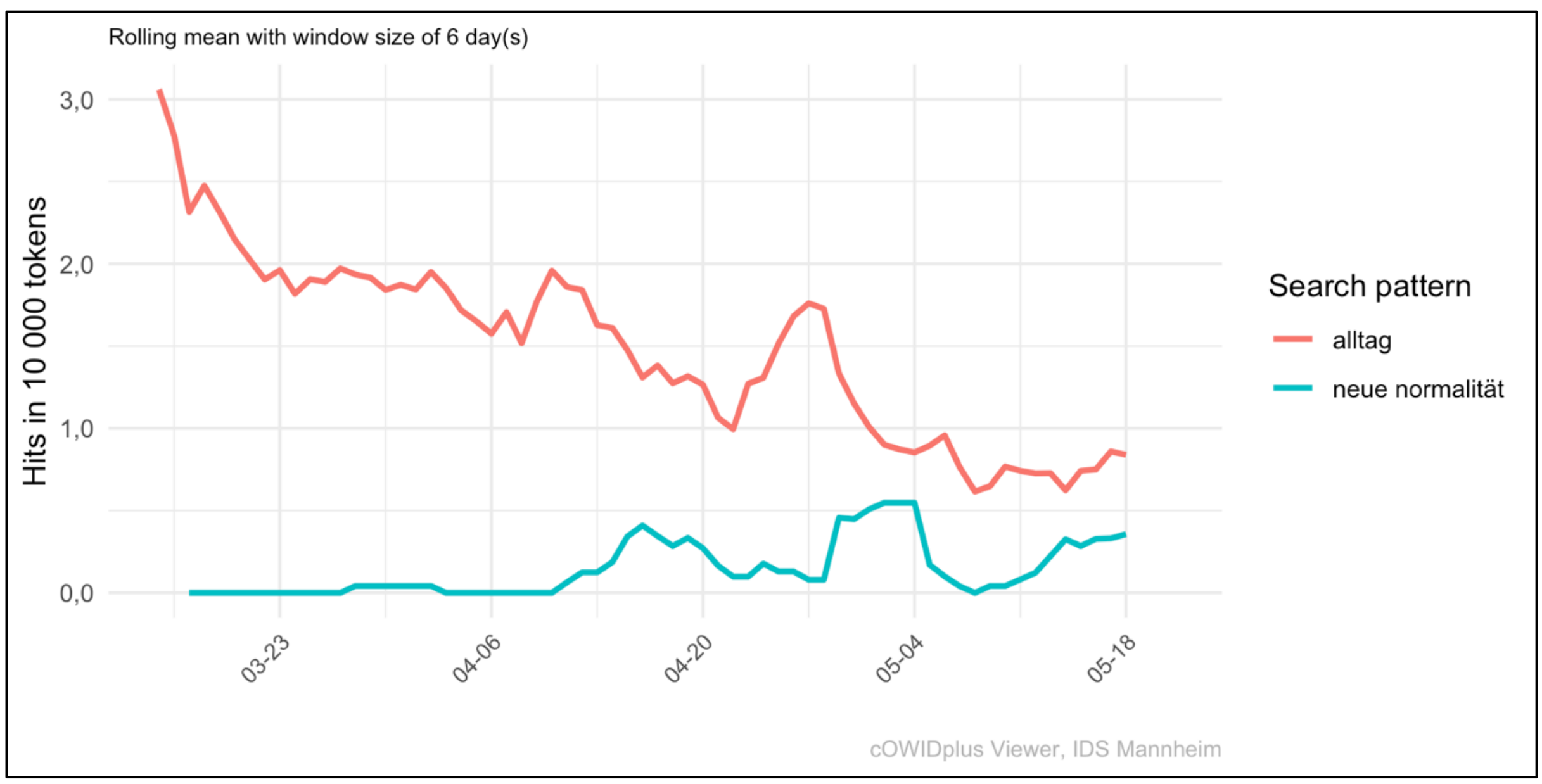

Figure 6: cOWIDplus Viewer result graph for exact matches of the unigram *alltag* (“everyday life”) and the bigram *neue normalität* (“new normality”) since March 15th.

The last example shows how unigram and bigram searches can be mixed. In Figure 6, relative frequencies for *alltag* (“every-day life”) and *neue normalität* (“new normality”) are shown. Although *alltag* remains more frequent throughout the whole time period, the German-speaking online-press begins to include references to a certain *neue normalität* into their RSS feeds around the middle of April. It is quite reasonable to assume that it needed some time after the pandemic arrived in Germany that such a “new normality” is being established and written about.

### 4.3 Scope and comparison

The two examples in the previous section show that people with little or no knowledge about data manipulation and/or statistical programming are able to track some of the effects the coronavirus pandemic had and has on the vocabulary in the German-language online press. However, as time passes and the pandemic (hopefully) becomes less severe, its influence on press discourse is likely to diminish. Of course, this does not mean that the viewer will lose its relevance. On the contrary, a resource which can be used with little knowledge about programming and statistics to track the most recent changes in online press vocabulary with data that is updated on a weekly basis can be very useful even in "normal" times.

A related resource that was recently released is "The Coronavirus Corpus" that is part of the English-Corpora.org suite of corpora (https://www.english-corpora.org/corona/). This corpus consists of English articles harvested from the web that deal with the coronavirus. We are looking forward to interesting comparisons with our German-language resource. Note, however, that we did not make any restrictions on which items from the RSS newsfeeds are incorporated in our corpus. On a more general level, the NOW corpus (https://www.english-corpora.org/now/) – of which the Coronavirus Corpus is a subset of – provides an excellent case in point for the assertion made above that resources that make it possible to track real time changes in a corpus, are of great value both in linguistics and in the digital humanities.

Another interesting corpus resource is currently being built at the Center for digital lexicography of the German Language (ZDL). The idea of the "coronakorpus" is to collect language data from relevant URLs from a range of publication sources in the German language (e.g., blogs, forums, tweets, and general information websites on the topic but also online newspaper articles). As for the "Coronavirus Corpus" mentioned above, the text has to deal with the coronavirus pandemic to be incorporated into the corpus.

## 5 Summary

We presented three inter-connected resources aiming to capture, analyze and explore recent German RSS news feed data since the beginning of 2020. The primary data source is the RSS corpus which is used both in cOWIDplus Analysis and the cOWIDplus Viewer. The former is a static HTML page that is updated on a weekly basis and seeks to analyze how the diversity of the vocabulary is developing in the light of the coronavirus crisis. The latter is an online application that enables both researchers and the broader public without data processing and analysis skills to explore the development of word forms through time.

As time passes and the impact of the coronavirus crisis on language presumably becomes much weaker than in March and April, we believe that the Viewer will still prove to be a valuable tool to explore a historical record of German press language during a global crisis. Also, it enables the research community to compare the effects of such a (hopefully) singular event like this global pandemic to other events with large-scale implications (e.g., the US presidential elections later in 2020). In addition, we argued that the Viewer can also be used to analyze developments in vocabulary as close to real-time as possible. One obvious expansion of the scope would be to include more RSS sources in the corpus. This would expand the coverage of German language sources but would also allow the construction of corpora for other languages.

**Address for correspondence**

Sascha Wolfer
Department for Lexical Studies
Leibniz Institute for the German Language
R5, 6-13
68161 Mannheim
Germany

wolfer@ids-mannheim.de

**Co-author information**

Alexander Koplenig
Department for Lexical Studies
Leibniz Institute for the German Language

Frank Michaelis
Department for Lexical Studies
Leibniz Institute for the German Language

Carolin Müller-Spitzer
Department for Lexical Studies
Leibniz Institute for the German Language